%% file: main.tex
\documentclass[runningheads]{llncs}

\usepackage{float}
\usepackage{tabularx}
\usepackage{multirow}
\usepackage{amsmath}
\usepackage{booktabs} 
\usepackage{caption} 
\usepackage{subcaption} 
\usepackage{graphicx}
\usepackage{pgfplots}
\usepackage[all]{nowidow}
\usepackage{caption}
\usepackage[utf8]{inputenc}
\usepackage{tikz}
\usetikzlibrary{er,positioning,bayesnet}
\usepackage{multicol}
\usepackage{algpseudocode,algorithm,algorithmicx}
\usepackage{hyperref}
\usepackage{xcolor}
\usepackage{wrapfig,lipsum,booktabs}
\usepackage{nth}
\usepackage{tablefootnote}
\usepackage[export]{adjustbox} 

\usepackage[inline]{enumitem} 

\definecolor{blue}{HTML}{1F77B4}
\definecolor{orange}{HTML}{FF7F0E}
\definecolor{green}{HTML}{2CA02C}

\pgfplotsset{compat=1.14}

\definecolor{gold}{rgb}{1.0, 0.0, 0.0}
\definecolor{silver}{rgb}{0.376, 0.66, 0.376}
\definecolor{bronze}{rgb}{0.0, 0.0, 1.0}

\begin{document}
\title{Illumination Estimation Challenge: experience of past two years}

\author{\uppercase{Egor Ershov}\inst{1} \and  \uppercase{Alex Savchik}\inst{1} \and  \uppercase{Ilya Semenkov}\inst{1}\inst{2} \and  \uppercase{Nikola Bani{\'{c}}}\inst{4} \and \uppercase{Karlo Ko{\v{s}}{\v{c}}evi{\'{c}}}\inst{5} \and  \uppercase{Marko Suba{\v{s}}i{\'{c}}}\inst{5} \and  \uppercase{Alexander Belokopytov}\inst{1} \and  \uppercase{Zhihao Li}\inst{7} \and \uppercase{Arseniy Terekhin}\inst{1} \and  \uppercase{Daria Senshina}\inst{1} \and  \uppercase{Artem Nikonorov} \inst{3} \and  \uppercase{Yanlin Qian}\inst{8} \and \uppercase{Marco Buzzelli}\inst{6} \and \uppercase{Riccardo Riva}\inst{6} \and \uppercase{Simone Bianco}\inst{6} \and \uppercase{Raimondo Schettini}\inst{6} \and  \uppercase{Sven Lon{\v{c}}ari{\'{c}}}\inst{5} \and \uppercase{Dmitry Nikolaev}\inst{1}}

\institute{Institute for Information Transmission Problems, RAS, Bol'shoi Karetnyi per. 19, Moscow, Russian Federation \\ \and
National Research University The Higher School of Economics, Myasnitskaya Ulitsa 20, 101000 Moscow, Russia \\ \and
Image Processing Systems Institute of RAS, Molodogvardeyskaya st., 151, Samara, Russia \and
Gideon Brothers, Radni{\v{c}}ka 177, 10000 Zagreb, Croatia \\ \and
Faculty of Electrical Engineering and Computing, University of Zagreb, Unska 3, 10000 Zagreb, Croatia \\ \and 
University of Milano -- Bicocca, Department of Informatics Systems and Communication, Viale Sarca, 336, Milan, Italy \\ \and
Nanjing University, Xianlin Road 163, Qixia District, Nanjing, Jiangsu Province, People's Republic of China \\ \and
Huawei MultiMedia Team
}
\authorrunning{Egor Ershov et al.}
\titlerunning{Illumination Estimation Challenge}

\maketitle 
\input{chapters/0_abstract}
\input{chapters/1_introduction}

\input{chapters/2_1st_challenge}

\input{chapters/3_challenge_description}
\input{chapters/6_discussion}
\input{chapters/8_acknowledgements}

\bibliographystyle{ama.bst}
\bibliography{main}

\end{document}

%% file: chapters/0_abstract.tex
\begin{abstract}

Illumination estimation is the essential step of computational color constancy, one of the core parts of various image processing pipelines of modern digital cameras.
Having an accurate and reliable illumination estimation is important for reducing the illumination influence on the image colors.
To motivate the generation of new ideas and the development of new algorithms in this field, the 2nd Illumination estimation challenge~(IEC\#2) was conducted.
The main advantage of testing a method on a challenge over testing in on some of the known datasets is the fact that the ground-truth illuminations for the challenge test images are unknown up until the results have been submitted, which prevents any potential hyperparameter tuning that may be biased.

The challenge had several tracks: general, indoor, and two-illuminant with each of them focusing on different parameters of the scenes.
Other main features of it are a new large dataset of images (about 5000) taken with the same camera sensor model, a manual markup accompanying each image, diverse content with scenes taken in numerous countries under a huge variety of illuminations extracted by using the SpyderCube calibration object, and a contest-like markup for the images from the Cube+ dataset that was used in IEC\#1.

This paper focuses on the description of the past two challenges, algorithms which won in each track, and the conclusions that were drawn based on the results obtained during the 1st and 2nd challenge that can be useful for similar future developments.


\end{abstract}

%% file: chapters/1_introduction.tex
\section{Introduction}
\label{sec:introduction}

\begin{figure}
    \centering
    \includegraphics[width = \textwidth]{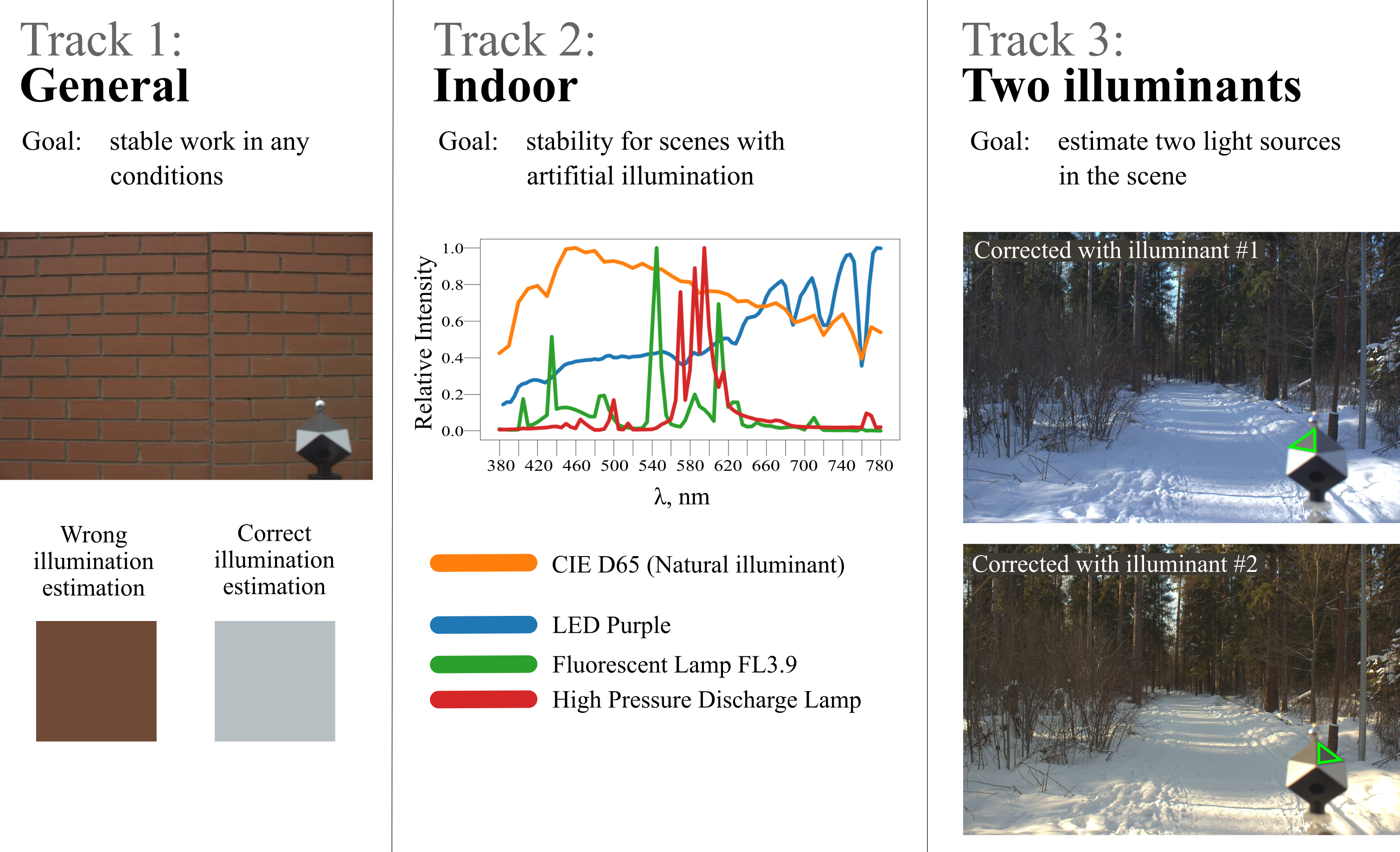}
    \caption{Ideas of 2nd Illumination Estimation Challenge illustration.}
    \label{fig:gr_abs}
\end{figure}

In the modern world of technology, a lot of modern devices, including mobile phones, tablets, and laptops are equipped with digital cameras. One of the essential parts of the image processing pipelines of these cameras is to remove the influence of the scene illumination on the image colors. To do that properly, it is first required to perform an accurate illumination estimation in an image or a video, and then this information is used for the actual color correction step~\cite{gijsenij2011color}.
Human color perception system performs a similar task by means of a feature known as color constancy~\cite{ebner2007color} that allows it to recognize object coloration regardless of the scene illumination. 
While the mechanism of color constancy in human color perception is not fully understood, there are numerous proposed implementations of the analogous feature in digital cameras known as auto white balance (AWB), which is supposed to achieve computational color constancy. 
Some of the simplest illumination estimation methods such as max-RGB~\cite{land1977retinex,funt2010rehabilitation,banic2014improving} or Gray-World~\cite{buchsbaum1980spatial} and its upgrades~\cite{finlayson2004shades,van2007edge} are based on simple image statistics and their main advantage are their speed and simplicity of implementation. 
Over time, the best accuracy used to be achieved by learning-based methods such as Bayesian learning~\cite{gehler2008bayesian}, spatio-spectral learning~\cite{chakrabarti2012color}, illumination solution space restriction~\cite{banic2015color,banic2015using,banic2015acolor}, using color moments~\cite{finlayson2013corrected}, regression trees with simple features from color distribution statistics~\cite{cheng2015effective}, spatial localizations~\cite{barron2015convolutional,barron2017fast}, convolutional neural networks~\cite{bianco2015color,shi2016deep,hu2017fc4,oh2017approaching}, genetic algorithms~\cite{koscevic2019color}, etc.
Thousands of other scientific papers about the problem of illumination estimation have been written, but it is still not solved to a satisfactory level due to the following reasons:
\begin{itemize}
    \item Volumes of publicly available scientific datasets are not large enough to cover all various cases of illumination. 
    \item Images in existing datasets have ground-truth scene illumination, but information about the scene is usually not provided. This excludes or complicates the process of a deeper study of the problem and its division into subtasks. For example, an efficient illumination estimation process can vary considering different times of the day. 
    \item Considering oversimplified formulation by evaluating only a single dominant light source (there are only few publicly available datasets with multiple light sources). The uniform illumination assumption is true for images taken on a cloudy day, or in a room with a single lamp. However, this assumption is not valid on a sunny day when there are at least two light sources: the sun and the sky. Another example includes a street at night and a closed room.
\end{itemize}

To take a step forward towards overcoming these three obstacles, two challenges for the task of illumination estimation were organized. The First Illumination Estimation Challenge (IEC\#1) was organized as a part of the \nth{11} International Symposium on Image and Signal Processing and Analysis (ISPA 2019, September 23-25, 2019, Dubrovnik, Croatia). A year later, the Second Illumination Estimation Challenge (IEC\#2) was organized as a part of the \nth{13} International Conference on Machine Vision (ICMV 2020, November 2-6, 2020, Rome, Italy). A large illumination estimation dataset was created as a key component of the challenges. The dataset contains around 5000 images captured with two Canon cameras (Canon 600D and Canon 550D), which share the same image sensor model. For each image, ground-truth illuminations with up to two light sources per image were extracted. Additionally, various information about the scene content was provided for each image. 

For both challenges, the ground-truth illuminations for the images from the test set were made publicly available only after the authors have submitted their solutions. This was done in order to prevent possible hyperparameter tuning that could compromise the testing fairness. Such testing with previously unknown ground-truth illumination can be seen as a potential advantage over testing on publicly available datasets.

The paper is structured as follows: Section~\ref{sec:iec_one} describes IEC\#1, Section~\ref{sec:challenge_description} describes IEC\#2, Section~\ref{sec:discussion} concludes the paper.



%% file: chapters/2_1st_challenge.tex
\section{The First Illumination Estimation Challenge}
\label{sec:iec_one}
As a new checkpoint in experimental results in the field of illumination estimation, the First Illumination estimation challenge (IEC1) was conducted as part of the \nth{11} International Symposium on Image and Signal Processing and Analysis (ISPA 2019) that was held in Dubrovnik, Croatia. The challenge gave the opportunity to researchers to benchmark new or existing illumination estimation methods on a newly created image dataset for which the ground-truth illumination was released only after the challenge. The goal was to check the performance of various methods when the ground-truth for the test set is not available immediately. The participants were assigned with the task to estimate a vector representing the color of the dominant illumination source in the new given dataset.

The challenge dataset was split into train and test parts. For method training, the participants were provided with the publicly available Cube+ dataset~\cite{banic2019unsupervised}, which includes both images and corresponding ground-truth illuminations. The test set consisted of $363$ images taken with the same camera that was used to capture images of the Cube+ dataset. The images in the test data were made public on the last day of the challenge by providing the password to decrypt the test archive that was already available earlier. The ground-truth illuminations corresponding to the images in the test set were made public in a similar manner after the challenge ended. By hiding the test set until the end of the challenge, it was intended to prevent or significantly reduce problems such as  potential data manipulation and method overfitting.

The quality of a solution was based on the reproduction angular error~\cite{finlayson2014reproduction}, which is defined as
\begin{equation}
    \label{eq:repr_error}
    R (\mathbf{g}, \mathbf{a}) = \arccos \left \langle \mathbf{1},  \frac{\mathbf{g} \oslash \mathbf{a}}{\left\lVert\mathbf{g} \oslash \mathbf{a}\right\rVert} \right \rangle,
\end{equation}
where $\mathbf{g}$ and $\mathbf{a}$ are three-dimensional vectors representing ground truth and estimated illumination, respectively, $\mathbf{1}$ is the vector of perfectly corrected white color, i.e. $\mathbf{1} = \frac{1}{\sqrt{3}} (1,1,1)$, $\oslash$ denotes element-wise division, $\langle \cdot \rangle$ denotes scalar product of two vectors, and $\left\lVert\cdot\right\rVert$ denotes the Euclidean norm.

Nine solutions were submitted and the final ranking is shown in Table~\ref{tab:iec1}. For the sake of transparency the submitted illumination estimations were also made available on the challenge website\footnote{\url{https://www.isispa.org/illumination-estimation-challenge/leaderboard}}\textsuperscript{,}\footnote{\url{https://archive.is/CE8WH}}. One of the main conclusions of the challenge was that using the median angular error may not be the best options. Namely, the worst angular errors are disregarded by the median. Because of that, some methods can be trained to specifically aim at achieving a good median error, while simultaneously not paying attention to the worst errors, which may be detrimental for an overall performance and the worst-case performance~\cite{savchik2019color}. This has lead to the conclusion that some other metric should be preferred over the median angular error.


\begin{table*}[h]
\normalsize
\caption{The results of the 1st Illumination Estimation Challenge ranked by the median of the reproduction angular errors calculated from the illumination estimations reported by the authors for the test images; the error statistics are reported in degrees ($^\circ$) and a lower error is equivalent to better estimation performance.}
\label{tab:iec1}
\centering
\begin{tabular}{|c|c|c|c|}

    \hline
    Authors & \textbf{Median} & Mean & Trimean\\
    \hline
    \hline
    
    \textcolor{gold}{\textbf{Alex Savchik, Egor Ershov, and Simon Karpenko}} & \textcolor{gold}{1.51} & \textcolor{gold}{2.65} & \textcolor{gold}{1.65}\\\hline
    
    \textcolor{silver}{\textbf{Jonathan T. Barron and Yun-Ta Tsai}} & \textcolor{silver}{1.59} & \textcolor{silver}{2.49} & \textcolor{silver}{1.73}\\\hline
    
    \textcolor{bronze}{\textbf{Yanlin Qian, Ke Chen, and Huanglin Yu}} & \textcolor{bronze}{1.64} & \textcolor{bronze}{2.93} & \textcolor{bronze}{1.77}\\\hline
    
    \textbf{Ke Chen, Huanglin Yu, and Yanlin Qian} & 1.69 & 2.61 & 1.84\\\hline
    
    \textbf{Yanlin Qian, Ke Chen, and Huanglin Yu} & 1.71 & 2.49 & 1.76\\\hline
    
    \textbf{Yanlin Qian, Ke Chen, and Huanglin Yu} & 2.10 & 6.87 & 2.50\\\hline
    
    \textbf{Viktor Vuk and V. N. Karazin} & 2.14 & 3.33 & 2.33\\\hline
    
    \textbf{Simon Karpenko, Egor Ershov, and Alex Savchik} & 4.58 & 6.68 & 5.07\\\hline
    
    \textbf{Hassan Ahmed Sial and Maria Vanrell I Martorell} & 5.91 & 7.29 & 6.18\\
    
    \hline

\end{tabular}
\end{table*}

%% file: chapters/3_challenge_description.tex
\section{The Second Illumination Estimation Challenge}
\label{sec:challenge_description}

\subsection{Dataset}
\label{subsec:ds_description}

For the challenge, the new dataset Cube++~\cite{ershov2020cube} was used. This dataset is the extension of the Cube+~\cite{banic2019unsupervised} dataset, which was used as a train set in IEC\#1. The Cube++ provides a lot of new images captured in various conditions, including indoor images and images with two illumination sources present in the scene. Such a dataset sets the ground for a broader range of illumination estimations research as indicated with multiple tracks of IEC\#2. To make the Cube++ compliant with its predecessor Cube+, the same camera sensor model and calibration object were used during the image acquisition. 

For ground-truth illumination extraction, the SpyderCube color target was used. Due to its cuboid shape, ground-truth illuminations can be extracted from two different directions. This is beneficial in two ways. First, by comparing such two ground-truths, it can be verified whether the scene illumination is uniform. Second, when needed, the ground-truth for two-illumination estimation can be extracted.

Along with the ground-truth illuminations, each image in the dataset is also accompanied by the metadata related to the conditions during image capturing procedure (ISO, exposure time, etc.), and manually labeled semantic information such as whether the image is captured indoor or outdoor, at what time of day was the image captured, is the image sharp, etc. Manually labeled semantic data was not available for the test set during the duration of the challenge.

The dataset contains files of three types: \texttt{name.jpg} - JPG images for preview; \texttt{name.png} - PNG images for illumination estimation; \texttt{name.json} - ground-truth illuminations and metadata.\footnote[1]{Detailed description of the dataset is available on GitHub \url{https://github.com/Visillect/CubePlusPlus} and IEC\#~2 website \url{http://chromaticity.iitp.ru}}

\subsection{Tracks}
\label{subsec:tracks}

\begin{wrapfigure}{R}{0.4\textwidth}
    \includegraphics[width=0.39\textwidth]{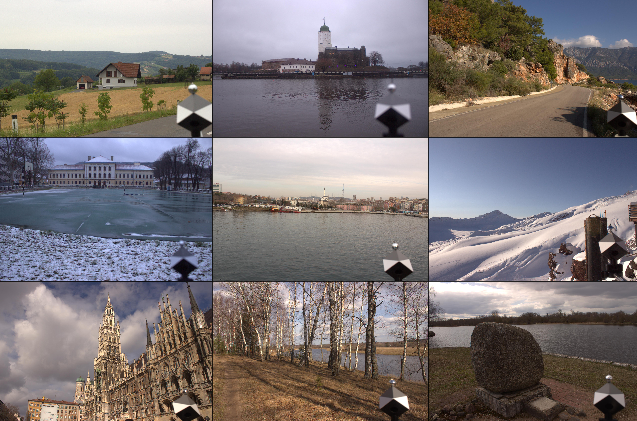}
    \caption{Examples of General track images.}
    \label{general_ex}
\end{wrapfigure}

The second illumination estimation challenge was organized in three tracks: general, indoor, and two-illuminant. Each track is described in more detail in the following text.

\subsubsection{General track.}
\label{subsubsec:general}

Usually, the quality of the illumination estimation algorithm is measured using a statistic of an error measure computed for all images in the test dataset. Very often, including in IEC\#1, the median of per image angular error values is used, which is reasonable in the case when the dataset itself has errors in markup.
Nevertheless, from a customer point of view, it is much more important not to have extremely wrong illumination estimations than to have the best values of only a single statistic.
Moreover, the problem of errors in markup should be solved not by metric selection, but by careful data labeling.
Therefore, the general track of the challenge was devoted to robust illumination estimation algorithms.

For the general track \textbf{dataset}, all images with angular difference less than $2^{\circ}$ between the ground-truth chromaticities extracted from left and right SpyderCube faces were selected from Cube++.
The dominant light source was chosen manually.

As a \textbf{metric}, the mean reproduction error described in~\eqref{eq:repr_error} and calculated for $25\%$ of the images with the worst error was used.

In Table~\ref{tab:general_leader}, the leaderboard for the general track is shown. The first place was achieved by Z. Li from Nanjing University with the CAUnet algorithm. This algorithm achieved the lowest mean reproduction error of $25\%$ of the worst estimations, which was $4.084077^{\circ}$.

\begin{table}[ht]
    \normalsize
    \caption{General track leaderboard}
    \label{tab:general_leader}
    \centering
    \makebox[0cm]{
    \begin{tabular}{|c|c|c|c|c|c|c|c|c|}
        \hline
        Team & Algorithm & \textbf{Mean (worst 25\%)} & Mean (worst 5\%) & Mean (worst 1\%) & Worst RE & Mean & Median & Trimean \\
    \hline
    \hline
    \textcolor{gold}{\textbf{Z. Li}} & \textcolor{gold}{\textbf{CAUnet}} & \textcolor{gold}{4.084077} & \textcolor{gold}{8.240} & \textcolor{gold}{13.313} & \textcolor{gold}{16.579} & \textcolor{gold}{1.605}  & \textcolor{gold}{0.966} & \textcolor{gold}{1.084} \\
    \hline
    \textbf{Z. Li} & \textbf{CAUnet} & 4.321251 & 8.270 & 13.064 & 20.294 & 1.725  & 1.084 & 1.207 \\
    \hline
    \textbf{X. Xing et al.} & \textbf{AL-AWB (sub. \# 2)} & 4.419051 & 8.421 & 13.299 & 19.995 & 1.822  & 1.197 & 1.317 \\
    \hline
    \textbf{X. Xing et al.} & \textbf{AL-AWB (sub. \# 1)} & 4.656795 & 8.851 & 13.338 & 19.995 & 1.891  & 1.230 & 1.352 \\
    \hline
    \textcolor{silver}{\textbf{Y. Qian}} & \textcolor{silver}{\textbf{sde-awb}} & \textcolor{silver}{4.979334} & \textcolor{silver}{10.283} & \textcolor{silver}{16.712} & \textcolor{silver}{21.626} & \textcolor{silver}{1.914} & \textcolor{silver}{1.164} & \textcolor{silver}{1.269} \\
    \hline
    \textbf{Y. Qian} & \textbf{sde-awb} & 5.112188 & 9.854 & 15.325 & 19.915 & 1.952 & 1.149 & 1.292 \\
    \hline
    \textbf{Y. Qian} & \textbf{sde-awb} &  5.377026 & 11.200 & 16.970 & 30.767 & 2.034 & 1.150 & 1.282\\
    \hline
    \textbf{J. Qiu et al.} & \textbf{illumGAN} & 9.999407 & 15.037 & 21.654 & 28.589 & 4.643 & 3.588 & 3.841 \\
    \hline
    BASELINE & \textbf{GreyWorld} & 10.419472 & 15.813 & 21.671 & 29.379 & 4.500 & 3.319 & 3.611 \\
    \hline
    BASELINE & \textbf{Constant} & 17.023748 & 27.369 & 35.637 & 40.972 & 7.081 & 4.020 & 5.275\\
    \hline
    \textcolor{bronze}{\textbf{J. Qiu et al.}} & \textcolor{bronze}{\textbf{illGAN1.0}} &  \textcolor{bronze}{25.954709} &  \textcolor{bronze}{32.458} &  \textcolor{bronze}{38.497} &  \textcolor{bronze}{40.849} & \textcolor{bronze}{19.325} & \textcolor{bronze}{18.278} & \textcolor{bronze}{18.499}\\
    \hline
    \textbf{J. Qiu et al.} & \textbf{illGAN1.0} & 26.058358 & 32.505 & 39.388 & 42.599 & 19.360 & 18.328 & 18.498 \\
    \hline
    \textbf{J. Qiu et al.} & \textbf{illGAN1.0} & 26.459121 & 32.924 & 39.715 & 42.599 & 19.408 & 18.468 & 18.710 \\
    \hline
    \end{tabular}
    }
\end{table}

\subsubsection{Indoor track.}
\label{subsubsec:indoor}

One of the stand-alone photography categories is indoor photography. 
Very often, under these conditions, the illumination in the scene is quite complicated: in different places of the scene there are many sources of illumination such as light from the window, incandescent lamps, LEDs, etc. 
Under these conditions, the determination of the dominant source in the scene may turn out to be a rather difficult task.

The indoor track \textbf{dataset} contains exclusively images for which it was manually determined that they are captured in indoor spaces. Additionally, the angle between ground-truth illuminations extracted from SpyderCube's left and right gray faces had to be less than $2^{\circ}$ to include an image into the dataset.

As a \textbf{metric}, the mean reproduction error described in~\eqref{eq:repr_error} was used.

In Table~\ref{tab:indoor_leader} the final indoor track ranking is shown. The first place was achieved by Y. Quian from Huawei Multiledia Team with an algorithm sde-awb which achieved the mean reproduction error of $2.541370^{\circ}$.

\begin{table*}[ht]
    \normalsize
    \caption{The results of the \nth{2} Illumination Estimation Challenge Indoor track ranked by the mean of the reproduction errors calculated from the illumination estimations reported by the authors for the test images; the error statistics are reported in degrees ($^{\circ}$) and a lower error is equivalent to better estimation performance.}
    \label{tab:indoor_leader}
    \centering
    \begin{tabular}{|c|c|c|c|c|c|c|}
    \hline
    Team & Algorithm & \textbf{Mean} & Median & Trimean & Mean (worst 5\%) & Worst RE \\
    \hline
    \hline
    \textbf{X. Xing et al.} & \textbf{AL-AWB (sub. \# 2)} & 2.500120 & 2.293 & 2.201 & 8.443 & 11.923 \\
    \hline
    \textcolor{gold}{\textbf{Y. Qian}} & \textcolor{gold}{\textbf{sde-awb}} & \textcolor{gold}{2.541370} & \textcolor{gold}{1.763} & \textcolor{gold}{1.943} & \textcolor{gold}{9.993} & \textcolor{gold}{10.976} \\
    \hline
    \textbf{X. Xing et al.} & \textbf{AL-AWB (sub. \# 1)} & 2.855412 & 2.293 & 2.407 & 10.954 & 13.665 \\
    \hline
    \textbf{J. Qiu et al.} & \textbf{illumGAN} & 3.191023 & 2.596 & 2.674 & 11.183 & 12.043 \\
    \hline
    \textcolor{silver}{\textbf{R. Riva et al.}} & \textcolor{silver}{\textbf{PCGAN, MCGAN}} & \textcolor{silver}{3.301088} & \textcolor{silver}{2.312} & \textcolor{silver}{2.298} &  \textcolor{silver}{16.861} & \textcolor{silver}{22.862} \\
    \hline
    \textbf{R. Riva et al.} & \textbf{PCGAN, MCGAN} & 3.376422 & 2.312 & 2.337 & 16.861 & 22.862 \\ 
    \hline
    BASELINE & \textbf{GreyWorld} & 4.105811 & 3.673 & 3.545 & 14.594 & 18.674 \\
    \hline
    BASELINE & \textbf{Constant} & 15.269933 & 14.802 & 15.332 & 29.241 & 29.996 \\ 
    \hline
    \end{tabular}
\end{table*}


\subsubsection{Two-illuminant track.}
\label{subsubsec:twoilluminant}

In everyday life, there are rarely situations where there is really only one source of illumination in the scene. 
Even during the daytime, it is customary to divide the illumination into two sources - the sun and sky. 
The main question of this track is whether it is possible to reliably extract more information about illumination using a single image. 
For these purposes, a dataset was assembled using a volumetric color target (SpyderCube) whose faces are illuminated by different sources.

Two-illuminant track \textbf{dataset} includes images for which the angle between ground-truth illuminations extracted from SpyderCube's left and right grey faces is greater than or equal to $2^{\circ}$. 
This helps to ensure that illuminations which are reflected from SpyderCube's faces are different enough to considered them originating from two different light sources.
Estimation of two light sources was required.

\begin{table*}[ht]
\normalsize
\caption{Two-illuminant track leaderboard}
\label{tab:twoilluminant_leader}
\centering
\makebox[0cm]{
\begin{tabular}{|c|c|c|c|c|c|}
    \hline
    Team & Algorithm & \textbf{Mean squared} & Mean & Median & Trimean \\
\hline
\hline
\textcolor{gold}{\textbf{Y. Qian}} & \textcolor{gold}{\textbf{sde-awb (sub. \# 1)}} & \textcolor{gold}{31.026217} & \textcolor{gold}{2.751} & \textcolor{gold}{2.262} & \textcolor{gold}{2.290}\\
\hline
\textbf{Y. Qian} & \textbf{sde-awb (sub. \# 2)} & 31.542930 & 2.737 & 2.171 & 2.309 \\ 
\hline
\textbf{X. Xing et al.} & \textbf{AL-AWB (sub. \# 2)} & 33.079119 & 2.657 & 1.844 & 2.082 \\
\hline
\textbf{Y. Liu et al.} & \textbf{3du-awb} & 37.305135 & 2.863 & 2.503 & 2.497 \\
\hline
\textbf{X. Xing et al.} & \textbf{AL-AWB (sub. \# 1)} & 41.883269 & 2.920 & 2.107 & 2.316 \\
\hline
BASELINE & \textbf{GreyWorld} & 81.840743 & 4.127 & 3.538 & 3.715 \\
\hline
BASELINE & \textbf{Constant} & 144.745182 & 5.264 & 3.475 & 3.815 \\ 
\hline
\end{tabular}
}
\end{table*}

As a \textbf{metric}, the squared sum of two angular reproduction errors was used:

\begin{equation}
    E(\mathbf{g}_1, \mathbf{g}_2, \mathbf{a}_1, \mathbf{a}_2) = 
    \min \Big (R^2 (\mathbf{g_1}, \mathbf{a_1}) + R^2 (\mathbf{g_2}, \mathbf{a_2}), R^2 (\mathbf{g_1}, \mathbf{a_2}) + R^2 (\mathbf{g_2}, \mathbf{a_1}) \Big),
\end{equation}
where $g_1$, $g_2$ are ground-truth chromaticity vectors and $a_1$, $a_2$ are algorithm estimations. Such strong metric allows to penalize algorithm with single answers for two close estimations.

\subsection{Challenge rules for all tracks}

\begin{wrapfigure}{r}{0.4\textwidth}
  \begin{center}
    \includegraphics[width=0.39\textwidth]{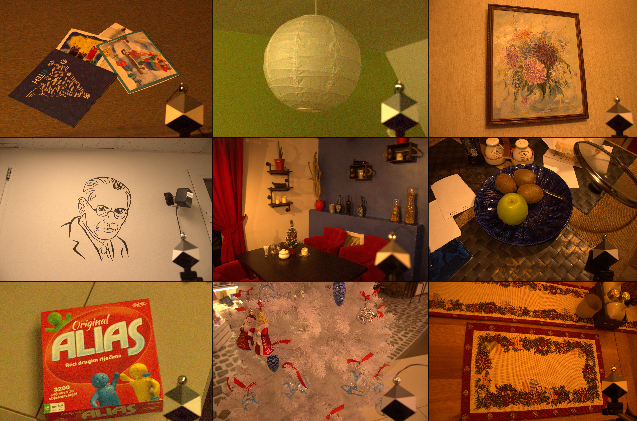}
  \end{center}
  \caption{Examples of Indoor track images.}
  \label{indoor_ex}
\end{wrapfigure}
Registration for the challenge was held up to 13:00 (GMT+3) on July 31, 2020. 
Participants were required to list both names and affiliations for all team members.

For each track, a corresponding dataset was prepared. 
Each team could have submitted up to 3 solutions for each track resulting in up to 9 submissions in total for 3 tracks.
Also, the code of metrics for final ranking was available at the official GitHub page. 
Participants were able to submit their solutions until 13:00 (GMT+3) on July 31, 2020. 
The test dataset was published a bit before submission day in an encrypted archive, the SpyderCube instance were masked out from the image scenes, the identifiers were shuffled to avoid finding similarities in close images with close identifiers, and there were no ground-truth illuminations nor manually annotated properties. 

\begin{wrapfigure}{R}{0.4\textwidth}
  \begin{center}
    \includegraphics[width=0.39\textwidth]{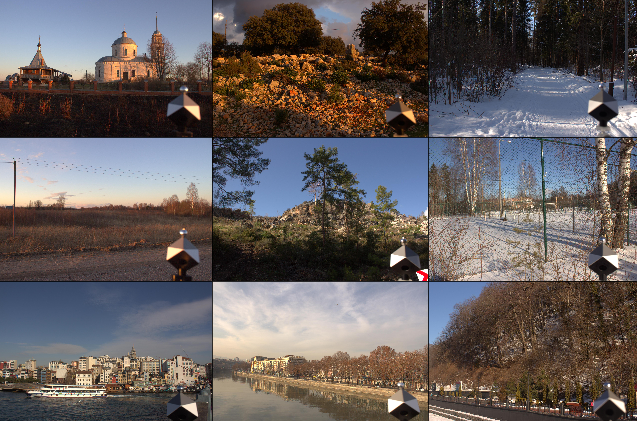}
  \end{center}
  \caption{Examples of Two-illuminant track images}
  \label{twoilluminant_ex}
\end{wrapfigure}

Finally, the key to the test part was published. 
The participants were required to run their model on the test part and to submit their prediction files. 

The leaderboard for participants was not available online until the final scoring to avoid overfitting to a test set.

The first place was achieved by Y. Qian (Huawei MultiMedia Team) with the sde-awb algorithm. 
The final squared sum of two angular reproduction errors was 31.026217 for this algorithm.

%% file: chapters/6_discussion.tex
\section{Discussion}
\label{sec:discussion}

The main goal for a participant in the described challenges is to create an algorithm that predicts the ground truth value for each image with the least error.
Thus, the best algorithm will be called the one that accurately predicts the answer proposed by the organizers.
However, would such an algorithm be the best for estimating illumination in a scene?
The answer to this question is decidedly negative.

Firstly, when using this type of datasets, the task is implicitly substituted from ``Estimate illumination parameters in the scene'' to ``Predict the color of the white patch in the scene''.
Provided that the color of a white flat object in the scene changes significantly with a change in its orientation, it turns out that the winning algorithm solves the problem of predicting the color of a white reference object in some way located better than others.
In other words, the winning algorithm is good at predicting one of the many averaged illumination estimation in the scene, but not all them and may be  not the best one.
The experience of the \textbf{two-illuminant track} suggests that even in simple daytime sun scenes, the difference in chromaticity between different sides of the cube can reach more than 10 degrees.
In the context of the above, the statement ``Algorithm X works with an accuracy of less than 1 degree'' becomes weaker, since the scatter of estimates of correct answers in some images can be ten times greater.

Secondly, you cannot write off errors in the markup.
In the first competition, the median of errors was used to combat this.
In this case, large values of errors on incorrectly labeled data have a lesser effect on the final characteristic of accuracy.
However, as shown by the results of the competition, such an error allows the solution to be forgiven for too many gross errors, which, combined with the specific structure of the dataset (more than half of the images were taken outdoors during the day), will lead to inadequate ranking of algorithms.
For this reason, in the last competition, much stricter metrics were used, and the data was marked up and processed with special attention.

These arguments lead us to a logical question: ``What is the best way to measure accuracy using such datasets?''
According to the organizers of the competition, this issue is not trivial and depends on the ultimate goal for which a technical solution is being developed.
In the case of the physical task of estimating illumination parameters in a scene, it is worth noting that the type of datasets used is not the most successful: even a three-dimensional achromatic calibration object (grey ball) allows collecting a very small amount of information about the lighting in the scene.
This is probably why in 2020, scientists from Simon Fraser University published a new dataset \cite{aghaei2020flying} in which the illumination for one scene was determined at once at many different points using flying drone.
In the tasks of forming digital photographs, where white balancing algorithms are still used, the stability of the algorithm is much more important than the average accuracy of its operation.
Thus, in this problem, it is largely more appropriate to look at the mean of the angular error over some percentage (1\%, 5\%, etc.) of the worst responses.
On the one hand, this kind of metrics still allows you to operate with angular values, on the other hand, to control the worst cases of the algorithm, which is critical from the point of view of the end user.

%% file: chapters/8_acknowledgements.tex
\section{Acknowledgement}
The work under data collection was supported by the Croatian Science Foundation under Project IP-06-2016-2092, and the work under data processing was supported by Russian Science Foundation under Grant 20-61-47089.